\documentclass[a4paper]{article}

\usepackage{INTERSPEECH2021}
\usepackage{multirow,enumitem,graphicx}
\usepackage{xcolor,url}

\newcommand\edited{\textsc{edited}}
\usepackage[normalem]{ulem}

\title{Assessing the Use of Prosody in \\ Constituency Parsing of Imperfect Transcripts}
\name{Trang Tran$^1$, Mari Ostendorf$^2$}
\address{
  $^1$University of Southern California, Institute for Creative Technologies\\
  $^2$University of Washington, Electrical and Computer Engineering}
\email{ttran@ict.usc.edu, ostendor@uw.edu}

\begin{document}

\maketitle
\begin{abstract}
  This work explores constituency parsing on automatically recognized transcripts of conversational speech. 
  The neural parser is based on a sentence encoder that leverages word vectors contextualized with prosodic features, jointly learning prosodic feature extraction with parsing. We assess the utility of the prosody in parsing on imperfect transcripts, i.e.\ transcripts with automatic speech recognition (ASR) errors, by applying the parser in an N-best reranking framework.
  In experiments on Switchboard, 
  we obtain 13-15\% of the oracle N-best gain relative to parsing the 1-best ASR output, with insignificant impact on word recognition error rate. 
  Prosody provides a significant part of the gain, and analyses suggest that it leads to more grammatical utterances via recovering function words. 

\end{abstract}
\noindent\textbf{Index Terms}: constituency parsing, spoken language, prosody

\section{Introduction}
\label{sec:intro}
Constituency parsing is
well studied on written text, including multilingual texts
\cite{Kitaev2019multilingual}, but work on parsing conversational speech is more limited, 
and parsers trained on written text do not work well on conversational speech. 
%
Early work in parsing conversational speech addressed challenges not present in written text, e.g.\ the lack of punctuation and the presence of disfluencies \cite{Charniak2001,Johnson2004}.
Later studies successfully incorporated prosodic features into parsing \cite{Kahn2005,Hale2006,Dreyer2007,Huang2010}, but the modest gains were outstripped by advances in neural architectures and contextualized word representations \cite{JamshidLou2019}. Additionally, most work used a prosody representation learned from  human-annotated prosodic features, e.g.\ ToBI \cite{Silverman1992}, which are expensive and require expert knowledge. 
Recent work with neural parsers  \cite{Tran2018,Tran2019} showed that automatically learned prosody representations can still be beneficial.
However, these studies were done on human transcripts, an unrealistic assumption for spoken language systems. 

A number of studies have leveraged parsing language models in an effort to improve automatic speech recognition (ASR), but research aimed at improving the parse of the output has been limited. One study \cite{Kahn2012} explored joint parsing and word recognition by re-ranking ASR hypotheses based on parse features, showing a reduction in word error rate (WER). Another study \cite{Marin2014} explored parsing in the context of domain adaptation and ASR name error detection. The authors showed that using output parse features improved re-scoring word confusion networks and benefited the detection of ASR errors and out-of-vocabulary regions. Recent work by \cite{Yoshikawa2016} studied joint parsing with disfluency detection on ASR transcripts, but they looked at dependency parsing and the method required extending the label set with speech-specific dependency type labels to handle mismatched words. All these studies only parsed transcript texts; prosodic features were not used.

The work presented here fills a gap by assessing the use of prosody in parsing ASR transcripts, where there is a question of whether ASR errors will lead to noisy acoustic-prosodic features.
We use a state-of-the-art neural parser combined with N-best hypothesis re-ranking, and confirm that prosody still provides a benefit in parsing conversational speech in experiments on the Switchboard corpus \cite{Godfrey1993}.
In addition, we provide qualitative analyses of where the approach provides the most gains and its effects on WER.
The approach leverages a simple integration of prosodic and lexical word vectors in a transformer encoder, which is a framework used in many language processing systems and thus is applicable to other tasks.

\section{Dataset and Metrics}
\label{sec:data-metrics}
The dataset in our work is Switchboard (SWBD) \cite{Godfrey1993}, a collection of spontaneous telephone speech between strangers prompted to talk about a specific set of topics. SWBD has been widely used for both parsing and ASR, and
to our knowledge, is the only large dataset of conversational English that has a corresponding parse treebank. 
We also assume known sentence boundaries, as in most prior work.
For training and tuning the parser, we use the transcripts from standard parsing data splits for train, development, and test sets (932, 144, 50 conversations), as in previous work on parsing SWBD, e.g.\ \cite{Charniak2001,Tran2019}.
For the re-ranking module (details in Section \ref{ssec:ranker}), we split the parsing development set into training and development \emph{subsets} with 75\%-25\% ratio, in which the sentences were randomly selected. This is because we are focusing on the effects of ASR errors on a pipelined system with a trained parser (with or without prosody), and we would not have parse hypotheses from the training data (since the parser has already seen these sentences).
The test set is the same as in parsing. 

Constituency parsing for written text is commonly evaluated using EVALB,\footnote{\url{https://nlp.cs.nyu.edu/evalb}} 
i.e.\ reporting F1 score on predicted constituent tuples $(l, a, b)$, where $l$ denotes the constituent label, and $a$ and $b$ denote the starting and ending indices of the constituent. However, this measure only works when the predicted parse and the reference parse have the same words. For evaluating parses on ASR transcripts, we use SParseval \cite{Roark2006}, a scoring program similar to EVALB but with mechanisms to account for ASR errors.\footnote{Our code is made publicly available at \url{https://github.com/trangham283/asr_preps}}
For bracket F1, SParseval requires an alignment between word sequences of the gold and predicted parses. We obtain this alignment with Gestalt pattern matching. SParseval also has the option to compute dependency F1, which does not require the word alignment, as this measure is based on head-percolated tuples of $(h, d, r)$ where $h$ is the head word, $d$ is the dependent, and $r$ is the relation between $h$ and $d$. We present F1 scores for both bracket (``brk'') and dependency (``dep'') F1.
The ``dep'' F1 scores are lower than the ``brk''  scores, because errors in word sequences directly contribute to lower recall.

\section{System Components}
\label{sec:methods}

\subsection{Automatic Speech Recognizer}
\label{ssec:asr}
We use an off-the-shelf ASR system, ASPiRE \cite{aspire}, which was trained on Fisher conversational speech data \cite{fisher}, available in Kaldi's \cite{kaldi} model suite. Briefly, the ASPiRE system was trained using a lattice-free maximum mutual information (LF-MMI) criterion, with computation efficiencies enabled by a phone-level language model and outputs at 1/3 the standard frame rate (one frame every 30 ms). The ASPiRE system has a reported word error rate (WER) of 15.6\% on the Hub5 `00 evaluation set.

ASR is run on Treebank sentence units, where the segmentation times are based on word times in the hand-corrected Mississippi State (MS-State) transcripts \cite{Deshmukh1998}, using an alignment of Treebank words to the MS transcript words.
For each sentence, we retain a set of (up to) 10 best ASR hypotheses. Shorter sentences often had fewer hypotheses; 62\% of the sentences have 9 or fewer hypotheses, 24\% have fewer than 5.
Word-level time alignments are a by-product of the ASR system.

\subsection{Parser}
\label{ssec:parser}
Our parser is composed of a multi-head self-attention (i.e.\ transformer \cite{Vaswani2017}) encoder and a span-based chart decoder proposed by \cite{Kitaev2018}, extended to integrate prosodic features as in \cite{Tran2019}. 
The parser takes as input a sequence of $T$ word-level features: $x_1, \cdots, x_{T}$. For each word $i$ in a sentence, the encoder maps input $x_i$ to a query vector $q_i$, a key vector $k_i$, and a value vector $v_i$, which are used to compute the labeled span scores $p(l, a, b)$. The chart decoder then learns to output the parse tree with the highest scores summed over all possible labeled spans.  

The input vectors $x_i = [e_i ; \phi_i ; s_i]$ are composed of word embeddings $e_i$, pause- and duration-based features $\phi_i$, and learned energy/pitch (E/f0) features $s_i$, which taken together represent a prosodically contextualized word vector. The word embeddings $e_i$ are pretrained BERT embeddings \cite{Devlin2019}, which have been shown to perform well on a variety of NLP tasks, and also to benefit parsing conversational speech transcripts despite the mismatch with written text \cite{JamshidLou2019,Tran2019}.
Pause- and duration-based features $\phi_i$ are composed of pause durations before and after each word; word durations are normalized by the mean duration of the word type in the training corpus.

The acoustic-prosodic features $s_i$ are learned via a convolutional neural network (CNN) from energy (E) and pitch (f0) contours as described in \cite{Tran2018}. Briefly, the frame-level energy and pitch features are extracted using Kaldi \cite{kaldi} and normalized for each speaker side of the whole SWBD conversations. The frames corresponding to each word are then extracted based on word-level time alignments. 
Each sequence of f0/E frames corresponding to a time-aligned word (and potentially its surrounding context) is convolved with $N$ filters of $m$ sizes (a total of $mN$ filters). The motivation for the multiple filter sizes is to enable the computation of features that capture information on different time scales. For each filter, we perform a 1-D convolution over the f0/E features with a stride of 1. Each filter output is max-pooled, resulting in $mN$-dimensional speech features $s_i$ for word $i$. These prosody representations are jointly learned with the parsing objective. 

\begin{figure}[hbpt]
\centering
\includegraphics[scale=0.6,trim=1cm 3.7cm 4cm 3.5cm]{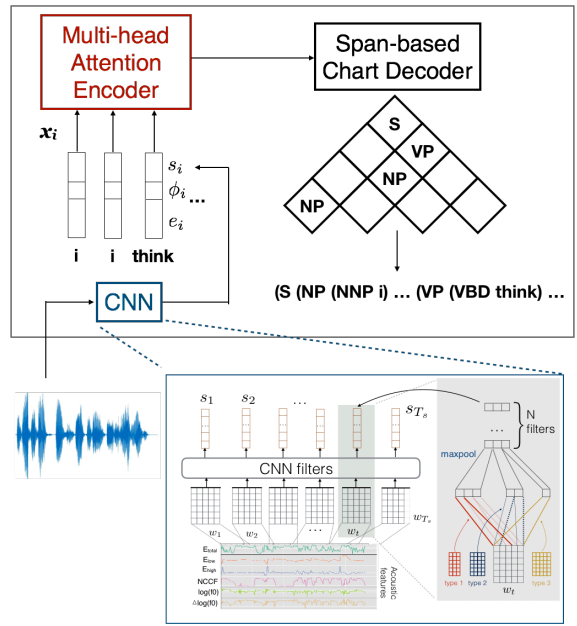}
\caption{Parser model overview, including: a CNN module for extracting prosodic features, a transformer encoder, and the chart decoder parser. Word-level input features include: word embeddings $e_i$, pause/duration features $\phi_i$, and acoustic-prosodic features $s_i$ learned by a CNN module.}
\label{fig:model-self-attn}
\end{figure}

Both parsers (with and without prosody features $\phi_i, s_i$) are trained on parses associated with hand transcribed speech, and hyperparameters are based on optimizing bracket F1 (from EVALB). Figure \ref{fig:model-self-attn} provides an overview of the parser and the CNN module.\footnote{We used the implementation provided at \url{https://github.com/trangham283/prosody_nlp/tree/master/code/self_attn_speech_parser}} 

\subsection{Ranker}
\label{ssec:ranker}
Given a set of ASR hypotheses for an utterance, we parse each hypothesis and train a ranker to select the hypothesis with the best F1 score. This process is formulated as a binary classification problem, e.g.\ as reviewed  in \cite{burges-overview}. Specifically, for each set of hypotheses, two sentences $a, b$ form a paired sample with features $F_{ab} = [f_{1a} - f_{1b}, \cdots, f_{Na} - f_{Nb}] $, where $f_{ix}$ is the $i$-th feature of a sentence $x\in\{ a,b\}$. These features include utterance length, number of disfluent nodes, parser output score, ASR output score, parse tree depth, total number of constituents in the predicted parse, and counts of several specific types of constituents such as \edited\ (disfluent nodes), \textsc{np}, \textsc{vp}, etc.
The prosodic cues are not used directly in ranking; they are implicitly included in the parser score.
The corresponding label $Y_{ab} = 1$ for the 
sentence pair if the F1 scores satisfy ${F1}(a)>{F1}(b)$; 
$Y_{ab} = 0$ otherwise. In constructing the training (sub)set, we select the pairs with the highest F1 score difference and 10 other random pairs. The ranker is the classifier $C(\cdot)$ that learns to predict $Y_{ab} = C(F_{ab})$. For each type of F1 score, i.e. ${F1}(\cdot) \in $ \{labeled, unlabeled\} $\times$ \{dependency, bracket\}, we trained a separate classifier/ranker to optimize for that score.

At test time, two ranking methods were used: point-wise and pair-wise. For point-wise ranking, each hypothesis sentence $a$ is considered individually to produce the probability score $P(a) = C(X_{a})$ (equivalent to a pairing of sentence $a$ with a sentence of all feature values 0). The best hypothesis is chosen by $\hat{a} = argmax_aP(a)$. For the pair-wise ranking method, two hypotheses are selected at a time, where the hypothesis for the next round of comparison is chosen based on its higher score. We report the results from the better ranking method in each setting. 

We experimented with several types of binary classifiers: logistic regression (LR), support vector machine classifier (SVC), and decision tree (DT). Hyperparameters of each classifier were tuned on the development (sub)set F1 scores. 
While more complex ranking approaches exist
(e.g.\ see \cite{burges-overview}), our feature set is small and the goal here is to demonstrate a benefit from prosody when considering multiple hypotheses. More complex ranking algorithms and/or the use of lattice ASR output are left for future work. 

\section{Experiments and Results}
\label{sec:exp-res}

\subsection{Ranking configuration}
\label{ssec:ranks}

The first set of experiments aimed at determining the best ranker and ranking features on the development set. Table \ref{tab:rankers-comp}
shows labeled dependency (``dep'') and bracket (``brk'') F1 scores on the development set, comparing different feature sets, parsing with only transcripts vs.\ transcript+prosody, and ranking classifiers. In almost all settings, the simple LR ranker outperforms SVC and DT (not shown, but results were similar to SVC), achieving the best dependency and bracket F1 scores of 0.520 and 0.713, respectively.

\begin{table}[hbpt]
    \centering
    \caption{\label{tab:rankers-comp}Labeled dependency (``dep'') and bracket (``brk'') F1 scores on the development set. `core set' denotes the feature set including: parser output score, ASR hypothesis score, sentence length, and number of \edited\ (disfluent) nodes. `depth' denotes parse tree depth; `$N_c$' denotes the total constituent count in the predicted parse; `*P' denotes the counts of several constituent types (\textsc{pp}, \textsc{np}, \textsc{vp}, \textsc{intj}) in the predicted parse.}
    \begin{tabular}{@{}clllll@{}}
    \toprule
    \multicolumn{1}{l}{} & Ranker & \multicolumn{2}{c}{LR} & \multicolumn{2}{c}{SVC} \\
    \multicolumn{1}{l}{} & feature set & \multicolumn{1}{c}{dep} & \multicolumn{1}{c}{brk} & \multicolumn{1}{c}{dep} & \multicolumn{1}{c}{brk} \\
    \midrule
    \parbox[t]{2mm}{\multirow{5}{*}{\rotatebox[origin=c]{90}{transcript}}} & core set & 0.514 & 0.701 & 0.513 & 0.699 \\
     & + depth & 0.513 & 0.699 & 0.513 & 0.697  \\
     & + $N_c$ & 0.513 & 0.700 & 0.512 & 0.698 \\
     & + depth + $N_c$ & 0.512 & 0.698 & 0.513 & 0.649 \\
     & + depth + $N_c$ + *P & 0.518 & 0.707 & 0.511 & 0.693 \\
     \midrule
    \parbox[t]{2mm}{\multirow{5}{*}{\rotatebox[origin=c]{90}{+prosody}}} & core set & 0.517  & 0.705 & 0.515 & 0.703 \\
     & + depth & 0.515 & 0.703 & 0.515 & 0.703 \\
     & + $N_c$ & 0.516 & 0.706 & 0.515 & 0.703 \\
     & + depth + $N_c$ & 0.513 & 0.706 & 0.515 & 0.704 \\
     & + depth + $N_c$ + *P & {\bf 0.520} & {\bf 0.713} & 0.512 & 0.697 \\ 
     \bottomrule
    \end{tabular}

\end{table}

Within the LR results, the best performing feature set consists of parse score (raw and normalized by length), ASR score (raw and normalized by length), sentence length, total number of constituents in the predicted parse, parse tree depth, and the number of certain types of constituents in the predicted parse: 
\edited, \textsc{intj}, \textsc{pp}, \textsc{vp}, and  \textsc{np}.
The parser trained with prosody features slightly outperforms the text-based one: 0.713 vs.\ 0.707 for bracket F1, and 0.520 vs.\ 0.518 for dependency F1. For the remaining results, we focus on this configuration: LR ranker with the full feature set. 

\subsection{ASR hypotheses vs.\ 1-best and the use of prosody}

Table \ref{tab:results-dev} presents results comparing the baseline (1-best hypothesis) with the best ranking strategy (LR ranker and full parse feature set), and results from ranking based on the parse score alone.
Using only the parse score is worse than using the 1-best ASR hypothesis, but re-ranking using parse features improves performance for both transcript-only (``trans.'') and transcript+prosody (``+pros.'') parsers, in all types of evaluations (labeled and unlabeled, dependency and bracket F1). 
Prosody contributes 30-40\% of the gain for the best case labeled F1 scores. 
For the bracket dependencies using reranking, the differences relative to the baseline and the difference when adding prosody are all statistically significant at $p<0.05$ using the bootstrap test \cite{efron93}. F1 scores for parsing on hand transcripts (``gold,'' i.e.\ best-case) range from 0.91-0.94, so there is still a large gap. Note that the gap in bracket F1 is smaller because the parser was tuned on the EVALB (bracket F1) objective, as is standard in parsing studies.

\begin{table}[hbpt]
\centering
\caption{\label{tab:results-dev}F1 scores on the development set across different sentence selection settings.}
\begin{tabular}{@{}llcccc@{}}
\toprule
 & selection by  & \multicolumn{2}{c}{unlabeled} & \multicolumn{2}{c}{labeled} \\ 
 & sentence's & dep & brk & dep & brk \\
 \midrule
 & 1-best ASR & 0.624 & 0.723 & 0.513 & 0.699 \\
 \midrule
\parbox[t]{2mm}{\multirow{3}{*}{\rotatebox[origin=c]{90}{trans.}}} & parse score & 0.588 & 0.698 & 0.499 & 0.664 \\
 & best ranker & 0.627 & 0.736 & 0.518 & 0.707  \\
 \cmidrule{2-6}
 & gold & 0.930 & 0.933 & 0.905 & 0.924 \\
\midrule
\parbox[t]{2mm}{\multirow{3}{*}{\rotatebox[origin=c]{90}{+pros.}}} & parse score & 0.594 & 0.706 & 0.502 & 0.670 \\
 & best ranker & {\bf 0.629} & {\bf 0.740} & {\bf 0.520} & {\bf 0.713} \\
  \cmidrule{2-6}
 & gold & 0.933 & 0.938 & 0.909 & 0.928\\
 \bottomrule
\end{tabular}
\end{table}


The results on the test set (Table \ref{tab:results-test}) confirm the findings that re-ranking benefits are greater for bracket scores and that parsers that use prosody consistently give better performance than those without prosody. In contrast to results on the dev set, the benefit from prosody on the test set is greater for labeled dependencies than for labeled brackets, and for dependencies it provides more than 70\% of the gain. 
The combination of re-ranking and prosody obtains 2-3\% relative improvement in F1 for the labeled cases, which corresponds to 13-15\% of the oracle possible gain with the N-best setting used here.

\begin{table}[hbpt]
\centering
\caption{\label{tab:results-test}Test set F1 scores compared between different systems. ``transcript'' and ``+prosody'' denote results after re-ranking outputs of the parser without and with prosody. ``oracle F1'' denotes results achieved by selecting best sentence-level F1 score in the set of hypotheses and ``gold'' denotes results on hand transcripts with the best parser (including prosody).
}
\begin{tabular}{lcccc}
\toprule
 & \multicolumn{2}{c}{unlabeled} & \multicolumn{2}{c}{labeled} \\
 & dep & brk & dep & brk \\ 
\midrule
1-best ASR & 0.612 & 0.700 & 0.491 & 0.676 \\
transcript & 0.619 & 0.714 & 0.494 & 0.687 \\
+prosody & 0.622 & 0.715 & 0.504 & 0.690 \\
\midrule
oracle F1 & 0.704 & 0.807 & 0.576 & 0.783 \\
gold & 0.934 & 0.933 & 0.909 & 0.926 \\
\bottomrule
\end{tabular}
\end{table}

SParseval by default does not include 
\edited\ 
(disfluent) nodes in scoring. This could be a disadvantage for our parser as it was trained to explicitly detect 
\edited\ nodes, so we also compute a modified SParseval score that considers 
\edited\ nodes.  Scores are generally lower when 
\edited\ nodes are included, but findings are similar except that the labeled dependency score benefits more from prosody. 


Direct comparison with previous work is difficult. Work by \cite{Marin2014} use a different dataset; \cite{Yoshikawa2016} use a different metric from SParseval; and \cite{Kahn2012} use parse scoring based on the whole turn instead of sentence units. 
Further, each of these works used a different ASR system to generate automatic transcripts, different ranking algorithms, and potentially different time alignments. 
With this caveat, 
the closest point of comparison is \cite{Kahn2012}, which reports results on Switchboard using an ASR system reporting 24.1\% 1-best WER (16.2\% N-best oracle WER, N=50) on the test set. Using reference sentence segmentations (similar to our scenario), they reported an unlabeled dependency F1 score of 0.734 with the oracle result of 0.823. The higher scores (despite the higher WER compared to our system) probably reflect differences in a scoring implementation that incorporates sentence segmentation.

\subsection{Effects on WER}

Table \ref{tab:wer-test} shows the test set WER with different parse ranking objectives using the best (transcript+prosody) parser. Excluding the oracle F1 case, the differences compared to the 1-best ASR hypothesis are not significant. 

\begin{table}[hbpt]
\centering
\caption{\label{tab:wer-test}WER on the SWBD test set for different parse ranking objectives. WER=0.19 for the 1-best baseline.}

\begin{tabular}{lcccc}
\toprule
 & \multicolumn{2}{c}{unlabeled} & \multicolumn{2}{c}{labeled} \\ 
\cmidrule{2-5}
score  & dep & brk & dep & brk \\
\midrule
transcript &  0.20 & 0.19 & 0.20 & 0.19 \\
+prosody  &  0.20 & 0.20 & 0.19 & 0.19 \\
oracle F1  & 0.16 & 0.17 & 0.17 & 0.16 \\
\bottomrule
\end{tabular}
\end{table}

For further analysis, we compare hypotheses selected by the best parser/re-ranker and the 1-best hypothesis. 
The best system overall results in a slightly higher WER, but gives small F1 improvements in sentences where all 10 hypotheses are available, which tend to be longer.
This result could be because most of the sentences are short (mean = 1.8--3 tokens) for those not producing all 10 hypotheses; only longer sentences (mean = 12.7 tokens) have full 10 hypotheses. 

In sentences where the prosody parser/re-ranker outperformed the 1-best hypothesis, 35\% of these are associated with better WER, and 23\% with worse WER. In both cases, the majority of words involved are function words: 82\% when WER improved, 77\% when WER degraded.

Some anecdotal (but common) examples are shown below; {\bf bold text} denotes words corrected by the prosody parser that were otherwise wrong (\sout{strike out text}) or missed in the 1-best hypothesis or the transcript-only (with re-ranking) parser. The better parser appears to favor grammatically correct sentences.

\begin{itemize}[nosep]
    \item i mean that 's better than george bush \sout{you} {\bf who} came out and said no
    \item {\bf do} you like rap music
    \item {\bf it 's} bigger than just the benefits
    \item \sout{learn} {\bf i learned} not necessarily be the center of attention
\end{itemize}

Finally, we considered whether human transcription error \cite{Tran2018,Zayats2019} could be a confounding factor. 
The Switchboard parses are based on sentence transcriptions that were later corrected, and 27\% of the test sentence have at least one transcription error, in which case the gold parse is less reliable.
Analysis in \cite{Tran2018} indicates that prosody appeared to hurt performance in the subset with errors, hypothesizing that errors in the reference parse might explain this.

Indeed, as Table \ref{tab:trans-err} shows, the bracket F1 score in sentences without transcription errors are higher both for the parser/re-ranker (0.707 vs.\ 0.660) and the 1-best hypothesis system (0.693 vs.\ 0.648). Similarly, the WER is lower in sentences without transcription errors both for the parser/re-ranker (0.181 vs.\ 0.237) and the 1-best hypotheses (0.169 vs.\ 0.235). Within 5854 test sentences, 1616 have at least one transcription error based on the MS-State corrections.

\begin{table}[hbpt]
\centering
\caption{F1 score and WER comparing sentences with and without transcription errors in the SWBD test set.}
\label{tab:trans-err}
\begin{tabular}{lcccc} \toprule 
 & \multicolumn{2}{c}{bracket F1} & \multicolumn{2}{c}{WER} \\
Sentences: & 1-best & \multicolumn{1}{c}{ranker} & 1-best & \multicolumn{1}{c}{ranker} \\ 
\midrule
with error & \multicolumn{1}{r}{0.648} & 0.660 & \multicolumn{1}{r}{0.235} & 0.237 \\
without error & \multicolumn{1}{r}{0.693} & 0.707 & \multicolumn{1}{r}{0.169} & 0.181 \\
\bottomrule
\end{tabular}
\end{table}

\section{Conclusion}
\label{sec:conclusion}
We present a study on parsing ASR transcripts with a neural parser that incorporates prosodic information. Our simple re-ranking framework using standard parse tree features and ASR scores
obtains 13-15\% of the oracle N-best gain relative to parsing the 1-best ASR output with no significant impact on WER.
Further gains may be obtained with simple extensions such as a larger N, different ranking algorithms, and integrated parsing and sentence segmentation.
The results also demonstrate how a pipelined system is impacted by ASR errors.
In all settings, parsing using prosodic features outperforms parsing with only text (transcript) information. When parsing improvement is observed, words involved in the hypothesis selection change are mostly function words (around 80\%). 

The parsing task is used here as a general means of exploring the impact of realistic inputs (ASR) in speech understanding.
Although many language processing systems today do not explicitly use parsers, parsing continues to be an active area of research, in part because it is useful for interpretability studies \cite{blevins2018,liu2019}. In addition, it serves as a good proxy for assessing contextualized word representations for a range of language understanding tasks \cite{miaschi2020,zanzotto2020-kermit}. 
The work here introduces a method for incorporating prosody into contextualization of word vectors, jointly learning the prosodic
representations in a transformer-based encoder, within a relatively standard encoder-decoder framework.
As such, the method can easily be transferred to other spoken language processing tasks.

\section{Acknowledgements}
We thank the reviewers for their helpful feedback. This work was funded in part by the NSF, grant IIS-1617176. 
Any opinions or recommendations expressed in this material are those of the authors and do not necessarily reflect the views of the NSF.

\bibliographystyle{IEEEtran}
\bibliography{refs}


\end{document}


\maketitle

\noindent The following descriptions are based on the Reproducibility Checklist.

\section*{Experimental Settings}
\begin{itemize}
    \item A clear description of the mathematical setting, algorithm, and/or model: this is in main paper, Section 3.
    \item Submission of a zip file containing source code, with specification of all dependencies, including external libraries, or a link to such resources (while still anonymized): we have included a zip file with our code; more details on experimental steps and preprocessing can be found in the README.md doc that comes with the code.
    \item Description of computing infrastructure used: we ran all our experiments on a 64-bit Linux CPU machine, Intel(R) Core(TM) i7-8700 CPU 3.20GHz. 
    \item Average runtime for each approach: it took approximately 15 minutes for each ranker's hyperparameter search per feature set. This applies for the Support Vector Machine classifier (SVC); for the Logistic Regression (LR) classifier, it took approximately 5 minutes per configuration.
    \item Number of parameters in each model: this depends on the features used, so these range from 4 to 12 weight parameters. The smallest feature set includes: parse score, ASR score, ASR length, and edit count. The largest features set includes: parse score, ASR score, raw parse score, normalized ASR score, ASR length, edit count, tree depth, total constituent count, INTJ count, NP count, VP count, and PP count. The best performing feature set includes parser score, ASR score, ASR length, tree depth, and count of the following nodes: NP, VP, PP, INTJ, EDITED. 
    \item Corresponding validation performance for each reported test result: all of the results are included in log files (folder log\_medians).
    \item Explanation of evaluation metrics used, with links to code: the code that implemented the metrics is included. Constituency parsing for written text is commonly evaluated with using evalb,\footnote{https://nlp.cs.nyu.edu/evalb/} i.e.\ reporting F1 score, the harmonic mean of precision and recall on predicted constituent tuples $(l, a, b)$, where $l$ denotes the constituent label, and $a$ and $b$ denote the starting and ending indices of the constituent. However, this measure only works when the predicted parse and the reference parse have the same words. For evaluating parses on ASR transcripts, we use SParseval \cite{Roark2006}, a scoring program similar to evalb but has mechanisms to take into account ASR errors. 

For bracket F1, SParseval requires an alignment between word sequences of the gold and predicted parses. We obtain this alignment with Gestalt pattern matching.\footnote{https://docs.python.org/3.6/library/difflib.html} SParseval also has the option to compute dependency F1, which does not require the reference and predicted sequences to have the same words, as this measure is based on head-percolated tuples of $(h, d, r)$ where $h$ is the head word, $d$ is the dependent, and $r$ is the relation between $h$ and $d$.
\end{itemize}

\section*{Experiments with Hyperparameter Search}
\begin{itemize}
    \item Bounds for each hyperparameter: we tuned the regularization parameter (C) in LR and SVC with grid search: $C \in$ [0.0005, 0.0001, 0.005, 0.001, 0.05, 0.01, 0.5, 0.1, 1.0, 5.0, 10.0, 50.0, 100.0]
    \item Hyperparameter configurations for best-performing models: $C=0.0001$ for LR and $C=10.0$ for SVC.
    \item Number of hyperparameter search trials: 5, for 5 different random seeds.
    \item The method of choosing hyperparameter values: the hyperparameters were chosen based on best performance on the validation set, based on the corresponding F1 score (also mentioned in Section 3 in the paper).
\end{itemize}

\section*{Datasets Used}
\begin{itemize}
    \item Relevant statistics such as number of examples: we described these in main paper, Section 2.
    \item Details of train/validation/test splits: we described this in main paper, Section 2 and 3.
    \item Explanation of any data that were excluded, and all pre-processing steps: we described this in the main paper, Section 2 and 3.
    \item A link to a downloadable version of the data: we used the standard Switchboard dataset, which can be obtained with an LDC license, as described in Section 2. 
    \item For new data collected, a complete description of the data collection process, such as instructions to annotators and methods for quality control: not applicable. 
\end{itemize}

\section*{Additional Results}
Table \ref{tab:results-test} shows the detailed results of parsing with and without prosody on the test set. Focusing on labeled bracket F1, the relative improvement from using a ranker over the 1-best hypothesis is 1.5\% for the best text parser, and 2\% for the prosody parser. Achievable improvement in relation to the gap between oracle F1 score (sentence selected by best F1 score), the prosody parser helps cover 12.4\% of the gap, compared to 9.8\% by the text-based parser. 

\begin{table}[hbpt]
\centering
\begin{tabular}{lcccc}
\toprule
 & \multicolumn{2}{c}{unlabeled} & \multicolumn{2}{c}{labeled} \\
 & dep & brk & dep & brk \\ 
\midrule
ASR & 0.612 & 0.700 & 0.491 & 0.676 \\
trans. & 0.619 & 0.714 & 0.494 & 0.687 \\
+pros & 0.622 & 0.715 & 0.504 & 0.690 \\
oracle F1 & 0.704 & 0.807 & 0.576 & 0.783 \\
\%$\Delta$,trans. & 1.1\% & 2.0\% & 0.7\% & 1.5\% \\
\%$\Delta$,+pros & 1.7\% & 2.2\% & 2.5\% & 2.0\% \\
\%$\updownarrow$,trans. & 7.2\% & 13.0\% & 4.0\% & 9.8\% \\
\%$\updownarrow$,+pros & 11.1\% & 14.2\% & 14.7\% & 12.4\% \\
\bottomrule
\end{tabular}
\caption{\label{tab:results-test}Test set F1 scores: $\Delta$ denotes the relative improvement of the system over the 1-best hypothesis; $\updownarrow$ denotes the gain achieved relative to the oracle score. ASR denotes 1-best hypothesis; ``oracle'' is based on sentence level F1 score}
\end{table}

\begin{table}[hbpt]
\centering
\begin{tabular}{lcccc} \toprule 
 & \multicolumn{2}{c}{bracket F1} & \multicolumn{2}{c}{WER} \\
Sents & 1-best & \multicolumn{1}{c}{ranker} & 1-best & \multicolumn{1}{c}{ranker} \\ 
\midrule
w/ error & \multicolumn{1}{r}{0.648} & 0.660 & \multicolumn{1}{r}{0.235} & 0.237 \\
w/o error & \multicolumn{1}{r}{0.693} & 0.707 & \multicolumn{1}{r}{0.169} & 0.181 \\
\bottomrule
\end{tabular}
\caption{\label{tab:trans-err}F1 score and WER comparing sentences with and without transcription errors.}
\end{table}

Table \ref{tab:trans-err} shows F1 score and WER comparing sentences with and without transcription errors, where ``corrected transcripts'' are those based on the Mississippi State (MS-State) University correction project \citep{Deshmukh1998}.
Within 5854 test sentences, 1616 have at least one transcription error based on the MS-State corrections. The bracket F1 score in sentences without transcription errors are higher both for for the parser-reranker (0.707 vs.\ 0.660) and the 1-best hypothesis system (0.693 vs.\ 0.648). Similarly, the WER lower in sentences without transcription errors.  

\begin{figure*}
     \centering
     \includegraphics[scale=0.5]{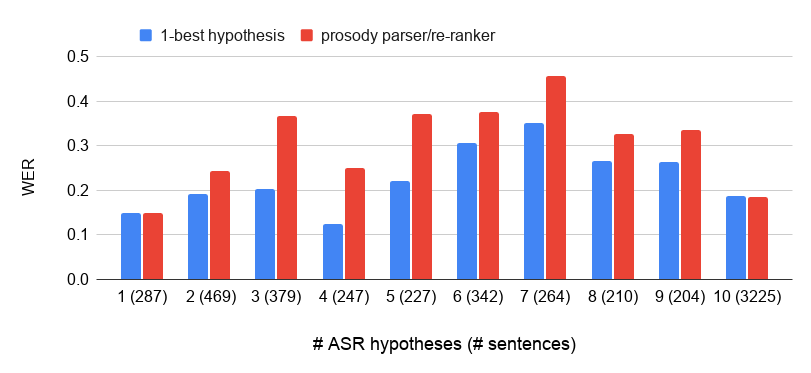}
     \caption{\label{fig:wer-test}WER on the test set grouped by the number of ASR hypotheses, comparing the baseline 1-best system and the best parser/re-ranker.}
\end{figure*}

Figure \ref{fig:wer-test} shows the word error rate on the test set grouped by the number of ASR hypotheses. The best system (transcript+prosody) parser-reranker results in better WER in short (single-hypothesis) sentences or where all 10 hypotheses are present.

\bibliographystyle{acl_natbib}
\bibliography{refs}
